\documentclass{article}

\usepackage[accepted]{icml2025}

\usepackage{times}
\usepackage{latexsym}

\usepackage[T1]{fontenc}
\usepackage{xspace}
\usepackage{hyperref}

\usepackage[utf8]{inputenc}
\usepackage{booktabs}
\usepackage{microtype}

\usepackage{inconsolata}

\usepackage{graphicx}

\newcommand{\eg}{\textit{e.g.},\xspace}

\icmltitlerunning{API Agents vs. GUI Agents: Divergence and Convergence}

\begin{document}

\twocolumn[
\icmltitle{API Agents vs. GUI Agents: Divergence and Convergence}

\begin{icmlauthorlist}
\icmlauthor{Chaoyun Zhang}{yyy}
\icmlauthor{Shilin He}{yyy}
\icmlauthor{Liqun Li}{yyy}
\icmlauthor{Si Qin}{yyy}
\icmlauthor{Yu Kang}{yyy}
\icmlauthor{Qingwei Lin}{yyy}
\icmlauthor{Saravan Rajmohan}{yyy}
\icmlauthor{Dongmei Zhang}{yyy}
\end{icmlauthorlist}

\icmlaffiliation{yyy}{Microsoft}

\icmlcorrespondingauthor{Chaoyun Zhang}{chaoyun.zhang@microsoft.com}

\icmlkeywords{GUI Agent, API Agent, LLM}

\vskip 0.3in
]

\printAffiliationsAndNotice{} 

\begin{abstract}
Large language models (LLMs) have evolved beyond simple text generation to power software agents that directly translate natural language commands into tangible actions. While API-based LLM agents initially rose to prominence for their robust automation capabilities  and seamless integration with programmatic endpoints, recent progress in multimodal LLM research has enabled GUI-based LLM agents that interact with graphical user interfaces in a human-like manner. Although these two paradigms share the goal of enabling LLM-driven task automation, they diverge significantly in architectural complexity, development workflows, and user interaction models.

This paper presents the first comprehensive comparative study of API-based and GUI-based LLM agents, systematically analyzing their divergence and potential convergence. We examine key dimensions and highlight scenarios in which hybrid approaches can harness their complementary strengths. By proposing clear decision criteria and illustrating practical use cases, we aim to guide practitioners and researchers in selecting, combining, or transitioning between these paradigms. Ultimately, we indicate that continuing innovations in LLM-based automation are poised to blur the lines between API- and GUI-driven agents, paving the way for more flexible, adaptive solutions in a wide range of real-world applications.
\end{abstract}

\section{Introduction}
The advent of large language models (LLMs) \cite{zhao2023survey} has ushered in a new era of artificial intelligence, enabling advanced natural language understanding and generation across a wide range of domains. While LLMs have long been recognized for their ability to produce coherent text, recent developments have led to LLM-based agents capable of mapping language inputs to real actions in digital environments \cite{wang2024survey}. By grounding LLMs in tangible operations, these agents can interact with various software systems, execute commands, and have a practical impact on the software ecosystems they inhabit.

Initially, software LLM agents were predominantly Application Programming Interface (API)-centric, interacting with external tools, functions, or services through well-defined programmatic interfaces \cite{du2024anytool}. This approach allowed agents to orchestrate microservices, query search engines, or even control third-party applications via documented APIs, with a automated and efficient manner. Products such as Microsoft's Copilot exemplify how API-based LLM agents have become mainstream \cite{stratton2024introduction}, rapidly transitioning from research prototypes to widely adopted industrial solutions. The traction of these agents, both in academia and industry, underscores their capacity to streamline tasks through automation while maintaining robust scalability and interoperability.

Concurrently, with multimodal capabilities gaining prominence in LLM research, a new class of agents has emerged: Graphical User interfaces (GUIs)-based LLM agents \cite{zhang2024large}. These agents interact not only through APIs but also by ``observing'' and manipulating graphical user interfaces of software, be they on desktop, mobile, or web applications. Projects such as UFO \cite{zhang2024ufo}, CogAgent \cite{hong2024cogagent}, and OpenAI Operator \cite{openai2025operator} illustrate how GUI-based agents can bring richer user experiences, improved accessibility, and provide more general automated control of software. By integrating vision or screen-based understanding with text-based reasoning, these agents push the boundaries of human-computer interaction, showcasing how AI can seamlessly blend with intuitive, visual workflows.

\begin{figure*}[t]
    \centering
    \includegraphics[width=0.7\textwidth]{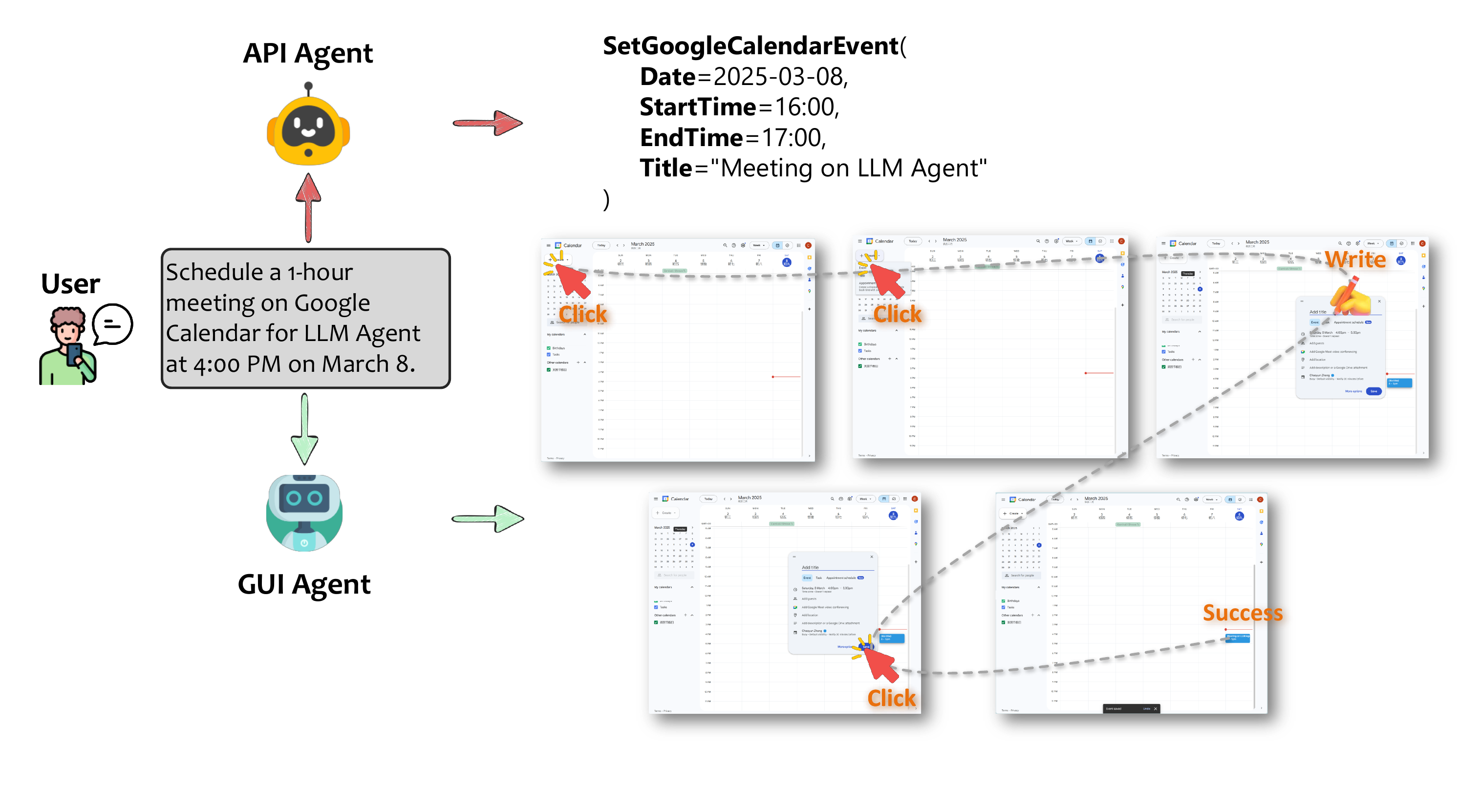}
    \vspace{-3em}
    \caption{The difference between an API agent and a GUI agent in completing the task ``Schedule a 1-hour meeting on Google Calendar for LLM Agent at 4:00 PM on March 8''.}
    \label{fig:vs}
    \vspace{-1.5em}
\end{figure*}

Despite the promise of both paradigms, API-based and GUI-based LLM agents exhibit significant differences in their architectures, development methodologies, and user interaction models \cite{zhang2024large}. For example, as illustrated in Figure~\ref{fig:vs}, an API-based agent scheduling a Google Calendar meeting could make a single call provided the proper endpoint and authentication, to instantly create the event. In contrast, a GUI-based agent would open the web interface, navigate through the calendar visually, fill in relevant fields, and click buttons to finalize the meeting details. While the API-based approach tends to be faster and more resource-efficient, it depends on well-defined endpoints and reliable infrastructure. Conversely, GUI-based agents can interface with application's front end but typically require multiple user-like steps, which can be slower and more error-prone \cite{song2024beyond}.

These trade-offs underscore the fundamental divergence between API- and GUI-based agents and have sparked vigorous discussions regarding their relative merits, overlap, and even necessity. Although they may appear to compete, many believe that a thorough examination of both differences and shared attributes is crucial. Indeed, no unified framework or comparative analysis has yet offered clear guidelines on when one approach might outperform the other—or if a hybrid strategy could combine their strengths.

To address this gap, this paper systematically examines both the divergence and potential convergence of API-based and GUI-based LLM agents in software automation. We begin by clarifying the conceptual foundations of each paradigm, followed by an in-depth comparative analysis across key dimensions. We then explore how ongoing innovations in LLM-based interactions are increasingly blurring the lines between API and GUI agents, and discuss a hybrid approach that leverages their respective strengths while mitigating their shortcomings. By outlining clear decision criteria, we aim to guide practitioners and researchers in selecting the most suitable agent type based on project requirements, resource constraints, and user experience objectives. Ultimately, this work serves as a comprehensive resource for academics and industry professionals, offering insights into how these agent types differ, where they converge, and how they may evolve to address emerging challenges.

\section{Background}
\begin{figure}[t]
    \centering
    \includegraphics[width=0.8\columnwidth]{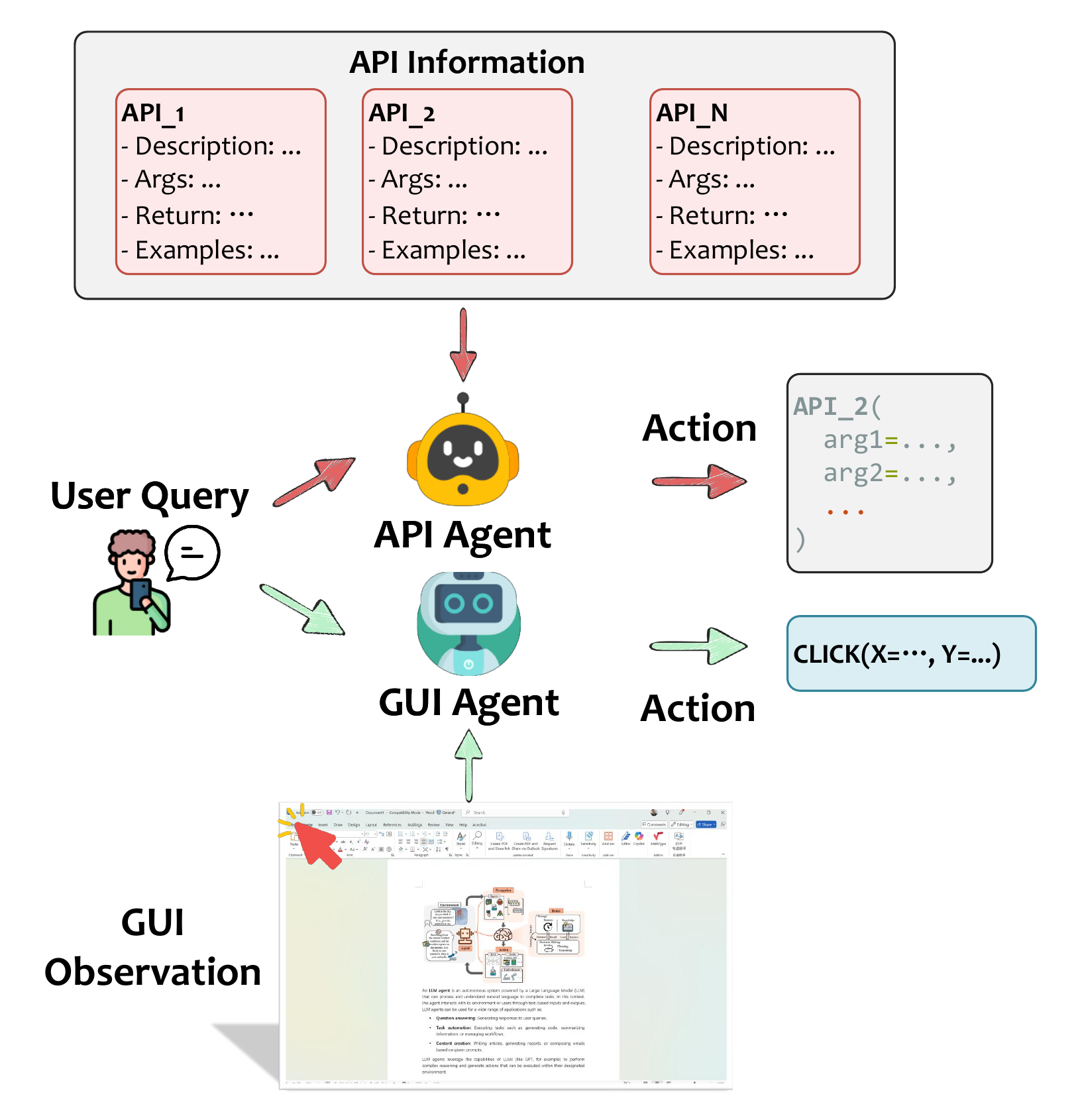}
    \vspace{-1em}
    \caption{The difference between an API agent and a GUI agent.}
    \label{fig:diff}
\end{figure}
LLM agents are designed to accept natural language requests from users and execute tasks in real or virtual environments. While they share the common goal of translating language into actionable commands, two distinct paradigms have emerged based on how these agents interact with their underlying systems: API-first and GUI-first LLM agents. Below, we provide an overview of each paradigm, highlighting their core principles and operational differences.

\subsection{API-Based LLM Agents}
An API-first LLM agent can be defined as
\begin{quote} 
    \textit{Intelligent agents that leverage LLMs as their cognitive engine to invoke one or more predefined APIs  to fulfill user requests automatically.} 
\end{quote}
In this paradigm, the agent's capabilities are bounded by a predefined set of tools, plugins, or function calls—collectively referred to as ``APIs'' \cite{shen2024llm}, as shown in Figure~\ref{fig:diff}. This constrained set of functions not only ensures reliability and safety but also simplifies the agent's decision space. Rather than generating full program code, the agent identifies which API to call for each step and populates the required parameters based on user intent.

Relevant API information, such as function names, descriptions, parameters, and schemas, is included in the LLMs prompt. When a user issues a natural language request, the LLM agent interprets the intent and chooses the most suitable API to perform the task. This approach has been widely adopted in many state-of-the-art models and frameworks, including function-call modes in GPT-4 \cite{hurst2024gpt} and the plugin-only mode in TaskWeaver \cite{qiao2023taskweaver}.

\subsection{GUI-Based LLM Agents}
In contrast to the API-first paradigm, GUI-based LLM agents operate by interacting directly with GUIs rather than invoking predefined functions. As defined in \cite{zhang2024large}, these agents can be described as:
\begin{quote} 
    \textit{Intelligent agents that operate within GUI environments, leveraging LLMs as their core inference and cognitive engine to generate, plan, and execute actions in a flexible and adaptive manner.} 
\end{quote}
GUI-based agents primarily rely on visual or multimodal inputs, such as application screenshots and textual representations (\eg accessibility trees or metadata). Instead of calling API endpoints, these agents navigate and manipulate on-screen elements using actions akin to human interactions, such as mouse clicks and keyboard inputs \cite{wang2024large}, as shown in Figure~\ref{fig:diff}. This ``human-like'' operational flow means that the agent must interpret visual layouts, locate relevant controls, and execute sequences of clicks, drags, or keypresses to accomplish tasks.

The rapid evolution of multimodal LLM \cite{hong2024cogagent, zheng2025vem} capabilities has sparked considerable interest in GUI-based agents, leading to a range of prototypes, frameworks, and commercial products. Use cases include automated software testing, workflow automation, and accessibility enhancements. Ongoing research aims to address the inherent challenges of screen parsing, error handling, and robust action planning, areas where GUI-based LLM agents continue to evolve.

Together with API-first LLM agents, this GUI-based paradigm forms a complementary, but often contrasting approach to building intelligent systems that bridge natural language understanding and real-world task execution. As subsequent sections will illustrate, understanding both paradigms is crucial for selecting or designing an agent architecture suited to specific project goals and constraints.


\section{Divergence Between API and GUI Agents}

\begin{table*}[t]
\centering
\caption{Comparison of API vs.\ GUI agents across key dimensions.}
\label{tab:divergence}
\begin{tabular}{p{2.8cm}| p{5.5cm}| p{5.5cm}}
\hline\hline
\textbf{Dimension} & \textbf{API Agents} & \textbf{GUI Agents} \\
\hline
Modality & Rely on text-based API calls & Depend on screenshots or accessibility trees \\\hline
Reliability & Generally higher with well-defined endpoints & Lower due to visual parsing and layout changes \\\hline
Efficiency & Achieve complex tasks in a single call & Require multiple user-like actions \\\hline
Availability & Limited to published or pre-defined APIs & Can operate on any visible UI element \\\hline
Flexibility & Constrained by existing APIs & Highly adaptable to new or unexposed features \\\hline
Security & Manageable via granular endpoint controls & Riskier due to broad access to UI elements \\\hline
Maintainability & Stable if APIs remain versioned & Prone to breakage on UI redesigns \\\hline
Transparency & Often hidden, back-end driven & Step-by-step, visually traceable \\\hline
Human-Like Interaction & Purely programmatic & Simulates user actions on a screen \\\hline
\hline
\end{tabular}
\end{table*}

Although both API-based and GUI-based agents aim to automate tasks using natural language instructions, they diverge significantly in various perspectives. This section presents a comparative analysis across several key dimensions: Modality, Reliability, Efficiency, Availability, Flexibility, Security, Transparency, Human-Like Interaction, and Maintainability. By examining these factors, we can better understand the foundational distinctions between the two paradigms and their impact on real-world applications.

\subsection{Modality}
The most evident difference lies in the way each agent perceives and interacts with software. API agents rely on textual specifications for each available endpoint. They interpret the user's request, map it to the relevant function, and provide the necessary parameters for execution. By contrast, GUI agents process visual or multimodal inputs, such as screenshots or accessibility trees, and then navigate and manipulate user interface elements. Since GUI agents operate on actual interface controls, accurate visual grounding and interpretation become essential \cite{lu2024omniparser}. While accessibility trees can offer some structured information, image-based comprehension remains central to the agent's interaction in a GUI environment.

\subsection{Efficiency}
Efficiency covers both the time and computational resources required to complete a task. API agents can generally handle complex tasks in a single function call, minimizing latency and reducing the inference costs. In contrast, GUI agents frequently must perform a series of user-like actions—opening menus, typing text, clicking buttons—to accomplish the same goal. Even routine operations, such as navigating through application panes, can require multiple steps. This user-level approach, while intuitive, can slow task execution and increase operational overhead compared to a well-designed API \cite{zhang2024large}. If completing a task takes significantly longer than it would for a human, it can limit GUI agents' practical applicability.

\subsection{Reliability}
Reliability often stems from the complexity of the underlying interaction model. API agents typically exhibit robust performance when they can access stable, well-defined endpoints. Such endpoints are easily maintained, versioned, and tested, leading to predictable outcomes. GUI agents, however, encounter challenges whenever application layouts or screen elements change unexpectedly. This is because GUIs are primarily designed for human interaction, often containing redundant elements and potential distractions that can disrupt automated workflows \cite{mckay2013ui}. These changes introduce uncertainties in the agent's visual parsing and planning, making GUI-based approaches more prone to errors. In addition, the multi-step decision-making process of GUI agents can compound errors at each step, ultimately reducing overall accuracy. As a result, GUI agents often require ongoing refinements and are still not as production-ready as their API-based counterparts in many scenarios.

\subsection{Availability}
Availability depends on how readily an agent can access the functionality necessary to fulfill a user's request. API agents are constrained by the endpoints or functions that developers have defined and exposed. If a desired feature is omitted, the agent cannot invoke it directly. This is particularly common in mobile applications, where developers often restrict external API access to maintain control over their private ecosystems.

Conversely, GUI agents can interact with virtually any application that presents a GUI, without requiring explicit API definitions. The universality of GUI-based interaction can be an advantage in environments where no formal APIs exposed, but it also demands more sophisticated interpretation and error handling to manage diverse or evolving UIs.

\subsection{Flexibility}
In addition to availability, flexibility denotes how easily the agent can adapt to new or modified use cases. API agents can call only the APIs that have been developed, documented, and integrated in advance. Expanding their functionality depends on creating and deploying additional endpoints. GUI agents, on the other hand, can theoretically operate on any visible element within an interface, thereby offering a higher degree of freedom. This freedom, however, requires advanced multimodal reasoning capabilities to locate and interact with UI objects consistently.

\subsection{Security}
Security plays a crucial role in deciding which paradigm to adopt. API agents typically offer more granular protection, as each endpoint can be individually secured with authentication, access control, or rate limiting. GUI agents may inadvertently access parts of the interface that perform privileged or destructive operations, raising the risk of unintended consequences. Since graphical interfaces are designed primarily for human users, enforcing comprehensive security policies on automated, mouse-and-keyboard-like interactions can be challenging. As a result, GUI-based agents may require additional safeguards to avoid unauthorized operations or misuse \cite{zhang2024ufo}.

\subsection{Maintainability}
An other dimension relates to how easily the agent's functionality can be maintained and updated over time. API agents benefit from versioned, standardized interfaces. As long as the underlying endpoints remain stable, the agent logic remains mostly intact. New APIs can be seamlessly integrated into the agent by simply adding their descriptions to the prompt, ensuring easy maintenance.

By contrast, GUI agents are highly susceptible to interface redesigns, pop-up windows, layout shifts, and element renaming or relocation \cite{zhang2024attacking}, all of which can break the automation if the GUI agent is unfamiliar with the change. This fragility can substantially increase the cost and frequency of maintenance.

\subsection{Transparency}
From a user's perspective, transparency refers to the clarity of observing how the agent fulfills a task. API agents often execute behind the scenes, providing limited visibility into the step-by-step process. Users typically see the final outcome without knowing which endpoints were invoked. 

In contrast, GUI agents replicate user-level interactions, making each click and text entry visible as it unfolds. Rather than operating invisibly in the background like API calls, GUI agents offer a visible and interactive execution process, allowing users to observe, intervene, or adjust the workflow as needed. This design can be particularly beneficial in scenarios requiring step-by-step verification, training simulations, or automation of tasks where visual confirmation is necessary. This improves the interpretability of the workflow, allowing human observers to track and validate the agent's progress more intuitively.

\subsection{Human-Like Interaction}
Closely linked to transparency is the notion of simulating human behavior. API agents employ a purely programmatic approach, executing function calls directly without mimicking user interactions. They are optimized for efficiency, reliability, and scalability, but they lack any visual or interactive representation of the task execution. In contrast, GUI agents replicate the exact steps a human user would take—navigating through menus, filling in forms, and interacting with interface elements in a natural, sequential manner. This human-like execution enhances interpretability, making it easier for users to follow and understand the agent's actions, thereby fostering trust and a more intuitive user experience. This introduces a novel paradigm for human-computer interaction by bridging AI automation with user-centric workflows \cite{lin2025carbon}.

\subsection{Summary}

\begin{table*}[t]
\centering
\caption{Examples of convergence paths in hybrid agent systems.}
\label{tab:hybrid}
\begin{tabular}{p{4.5cm} | p{5.5cm} | p{5cm}}
\hline\hline
\textbf{Approach} & \textbf{Key Benefit} & \textbf{Primary Challenge} \\
\hline
API Wrappers Over GUI Tools & Provides a quasi-API experience for GUI-only software & Still relies on underlying GUI elements that may change \\\hline
Unified Orchestration Tools & Hides agent-type details from the user & Complex logic to choose between API and GUI in real time \\\hline
Low-Code / No-Code Solutions & Simplifies design of advanced workflows & May introduce hidden dependencies and abstractions \\
\hline\hline
\end{tabular}
\end{table*}

In summary, these dimensions illustrate the fundamental ways in which API-based and GUI-based agents diverge in practice. API agents provide efficiency, security, and reliability when backed by robust endpoints, but they are bounded by the limited set of exposed functions. GUI agents offer broad applicability and user-like workflows, yet they must overcome challenges in visual parsing, interface changes, and slower task execution. As the complexity of software ecosystems grows, understanding these divergent properties is essential for selecting the most suitable approach—or for designing hybrid solutions that combine the best attributes of both paradigms.

\section{Convergence and the ``Hybrid'' Approach}
\label{sec:hybrid}

Although API-based and GUI-based agents have traditionally been studied as separate paradigms, their conceptual foundations are not mutually exclusive. In practice, there are numerous scenarios where these approaches intersect or complement one another, leading to an emerging ``hybrid'' model. By drawing on the strengths of both API and GUI paradigms, this hybrid approach can achieve broader coverage of use cases, higher efficiency, and a more human-like interaction style. While still in its early stages, the potential for these convergent solutions to reshape how agents operate is becoming increasingly evident.

\subsection{API Wrappers Over GUI Workflow}

\begin{figure}[t]
    \centering
    \includegraphics[width=\columnwidth]{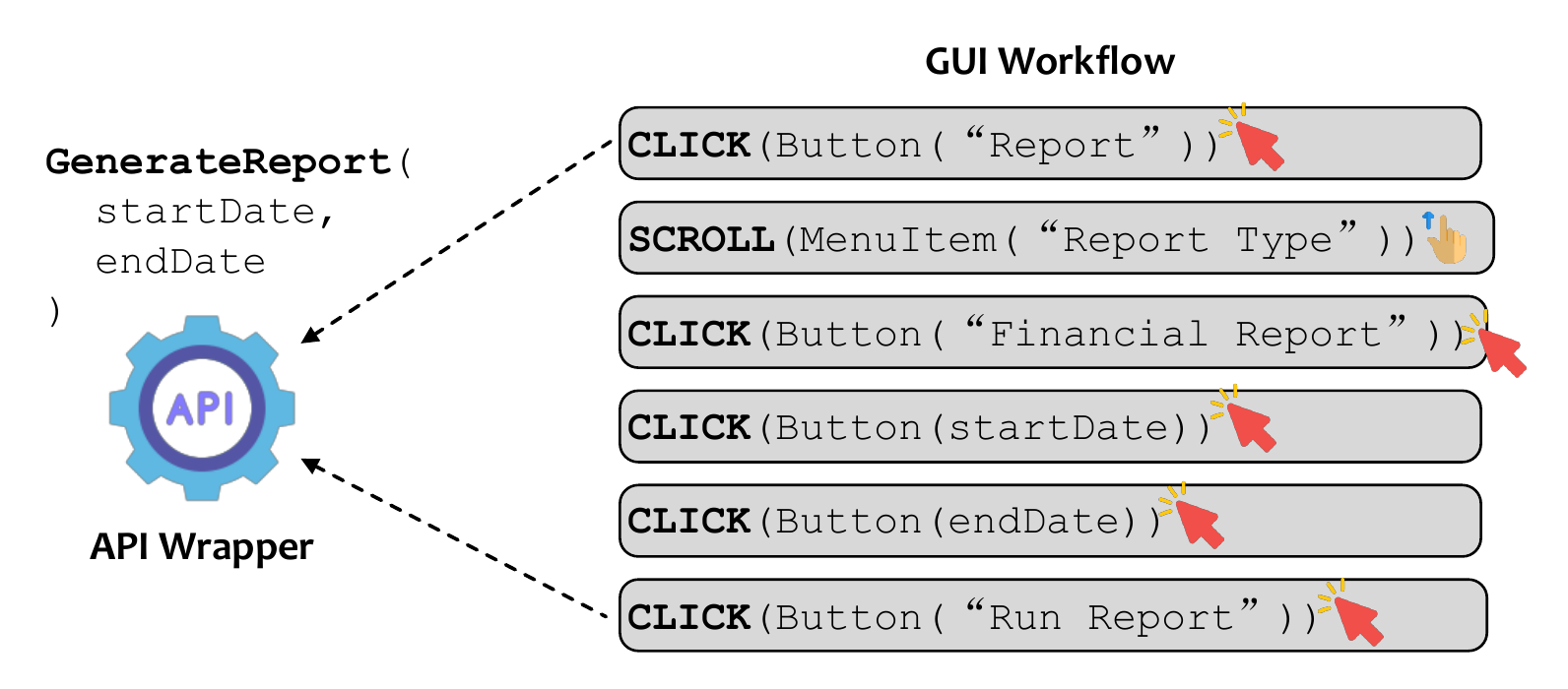}
    \vspace{-2em}
    \caption{An example of a API wapper over a GUI workflow.}
    \label{fig:wrapper}
    \vspace{-2em}
\end{figure}
Some vendors transform GUI-based applications into quasi-API services by introducing a ``headless mode'' or a scripting interface that accepts function calls. This approach effectively abstracts GUI interactions into structured commands, allowing applications originally designed for human navigation to be automated in a more programmatic and scalable manner. For example, a specialized accounting application may traditionally require users to navigate through multiple dialog boxes and menus to generate a financial report. However, in a headless or scripted version, the same application could expose a function such as \texttt{GenerateReport(startDate, endDate)}, enabling direct execution without requiring manual UI navigation. This is conceptually similar to Robotic Process Automation (RPA) bots \cite{wornow2024automating}, which mimic user actions but can also be optimized for backend workflows.

Although these wrappers still rely on GUI workflows under the hood, they present an API-like interface to developers, simplifying integration into broader automation pipelines. This transformation represents a subtle yet impactful form of convergence: an application originally designed for direct GUI interactions is reinterpreted as an API service, reducing the need for a dedicated GUI agent while retaining compatibility with existing software ecosystems. By bridging GUI automation with structured API-like interfaces, this approach enhances efficiency, scalability, and ease of integration, particularly in enterprise environments where legacy applications must be incorporated into modern automation frameworks.

\subsection{Unified Orchestration Tools}

\begin{figure}[t]
    \centering
    \includegraphics[width=\columnwidth]{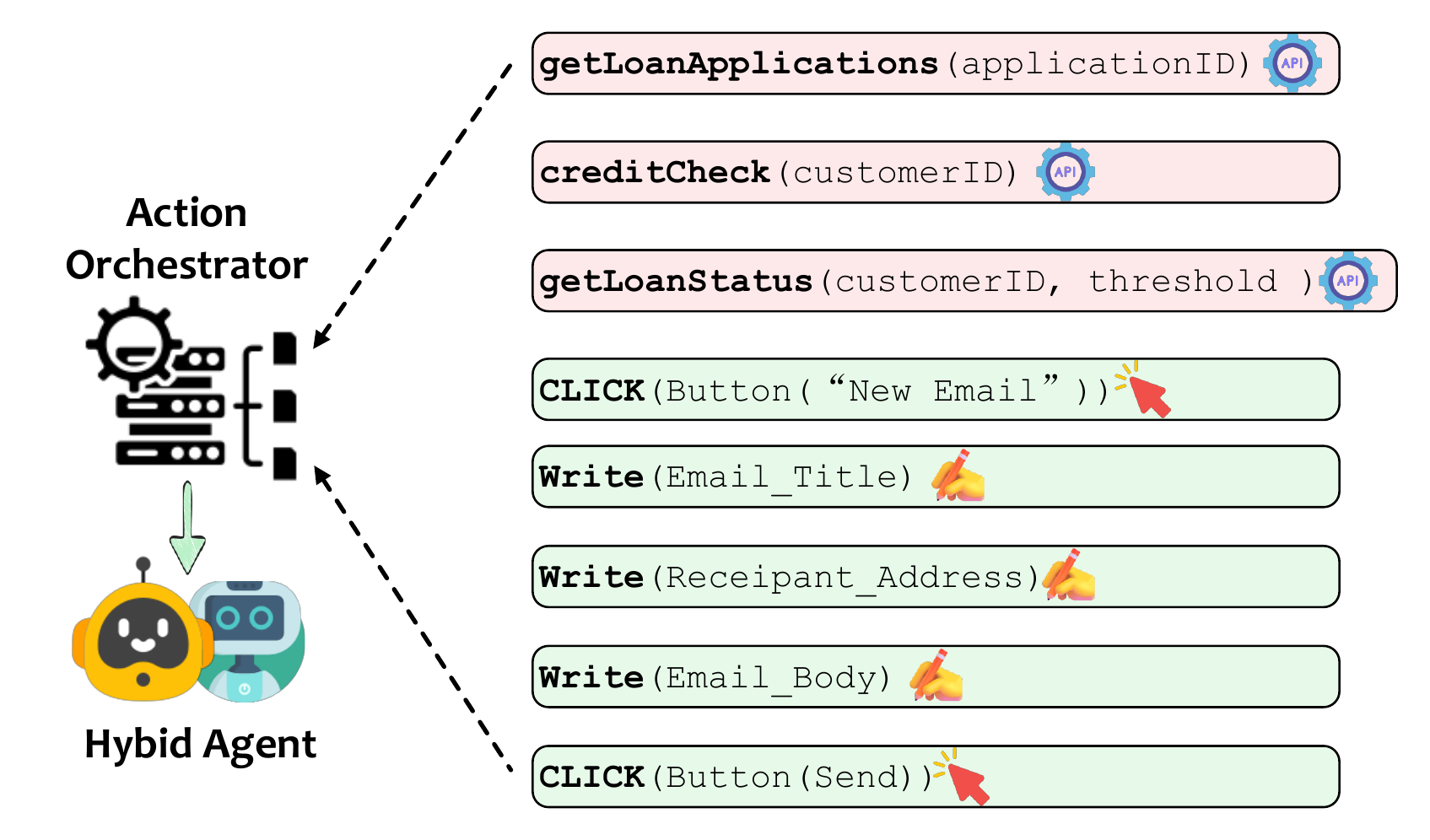}
    \vspace{-2em}
    \caption{An example orchestrator to manage API and GUI actions.}
    \label{fig:orchestration}
    \vspace{-1em}
\end{figure}
Enterprise-grade automation frameworks and process orchestration tools increasingly offer a single, unified environment where developers or operators can build high-level workflows without delving into the underlying agent mechanisms. Consider, for example, a large financial institution automating its loan approval process. Within an orchestration tool, a user could design a flowchart that checks a customer's credit score (using a secure API endpoint), then updates a Customer relationship management (CRM) system if the credit score meets a certain threshold. If no relevant API exists for updating the CRM, the platform can seamlessly switch to a GUI-based agent that navigates through the CRM's web interface in a user-like fashion. Thus, the tool automatically determines whether an API call or GUI interaction is most suitable for each task. We show such an example in Figure~\ref{fig:orchestration}. By shielding users from these low-level decisions, orchestration platforms reduce complexity and streamline the automation pipelines.

\begin{table}[t]
\caption{Comparison of Success Rate (SR) and Average Completion Steps (ACS) between GUI-only and GUI with API actions.}
\label{tab:api_compare}
\centering
\begin{tabular}{l|c|c}
\toprule
\textbf{Model} & \textbf{SR} & \textbf{ACS} \\
\midrule
GUI-only (GPT-4o)     & 16.3\% & 13.8 \\
GUI + API (GPT-4o)    & \textbf{22.4\%} & \textbf{12.9} \\
\midrule
GUI-only (o1)         & 16.3\% & 16.0 \\
GUI + API (o1)        & \textbf{24.5\%} & \textbf{6.6} \\
\bottomrule
\end{tabular}
\vspace{-2em}
\end{table}

\paragraph{Prototyping Experiments.}
We extend the UFO framework~\cite{zhang2024ufo} to prototype a hybrid GUI and API approach that prioritizes API usage whenever available, using GPT-4o and o1 as the foundation model. To evaluate its effectiveness, we focus on \textit{27} office-related tasks in OSWorld \cite{xie2024osworld} and manually implement \textit{12} APIs for Word, Excel, and PowerPoint. These applications expose COM interfaces that support the creation of custom functions, making them ideal candidates for deeper integration with the operating system and application layers. Table~\ref{tab:office_apis} summarizes the developed APIs in the Appendix~\ref{sec:office_apis}. 

Table~\ref{tab:api_compare} compares two configurations—GUI only and GUI with API—on two key metrics: \textit{(i)} overall Success Rate (SR) and \textit{(ii)} Average Completion Steps (ACS). To ensure fairness, ACS is computed only on the subset of tasks that both configurations complete successfully.

The results show that incorporating APIs improves SR for both GPT-4o (an increase of 6.1\%) and o1 (an increase of 8.2\%), demonstrating the benefit of combining GUI and API actions. GPT-4o benefits primarily by avoiding control detection failures, which often arise from unannotated interface elements. In contrast, o1 more frequently addresses planning errors by using direct API calls as shortcuts, reflecting its stronger reasoning capabilities and preference for concise execution plans.

In addition, the hybrid configuration reduces the effort required for task completion. Compared to GUI-only execution, the GUI with API setup achieves a 6.5\% reduction in steps for GPT-4o and a substantial 58.5\% reduction for o1. This improvement for o1 is due to its ability to strategically use APIs to bypass multiple GUI interactions. Overall, these confirm that combining GUI automation with API calls enhances both robustness and efficiency.


\subsection{Low-Code / No-Code Solutions}
\begin{figure}[t]
    \centering
    \includegraphics[width=\columnwidth]{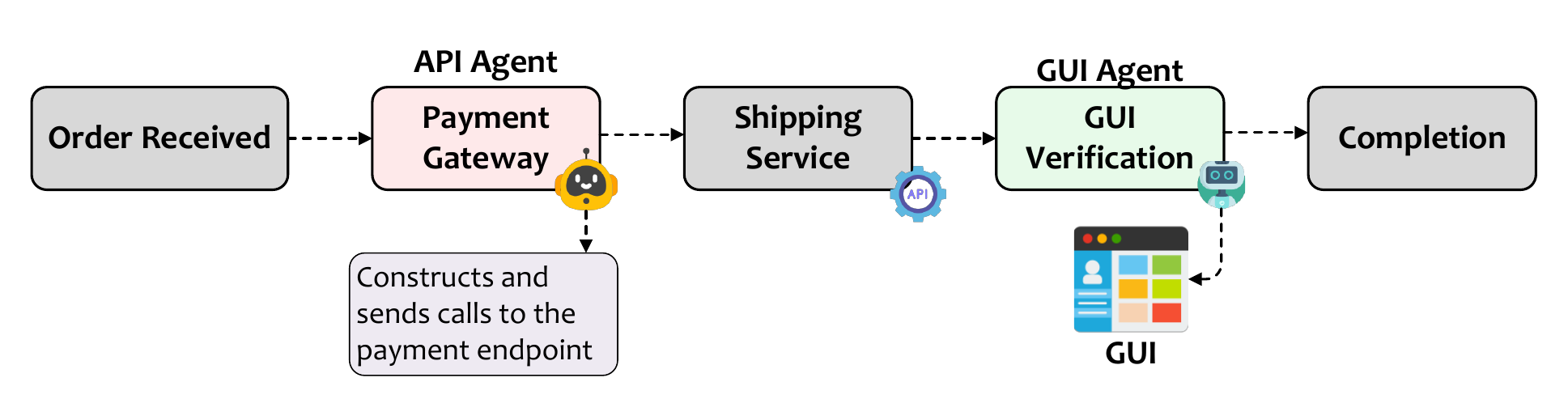}
    \vspace{-2.5em}
    \caption{One example of a no-code platform to create workflows integrating both API calls and GUI agents.}
    \label{fig:nocode}
    \vspace{-1.5em}
\end{figure}
Low-code and no-code platforms abstract many technical details behind visual interfaces, enabling non-experts (often referred to as ``citizen developers'') to construct applications or automations through drag-and-drop components \cite{tang2025metachain}. As illustrated in Figure~\ref{fig:nocode}, each block in this order-processing workflow represents a distinct task: the user drags a ``Payment Gateway'' block into the designer to handle transactions, configures it visually,while behind the scenes an API agent automatically constructs and sends calls to the payment endpoint. It then connects to a ``Shipping Service'' block for fulfillment. Behind the scenes, the platform typically issues API calls to the payment and shipping services, allowing the user to focus on the workflow’s logical sequence rather than on lower-level protocols. Conversely, if a given step calls for GUI-based verification—for instance, checking a specific user interface element on a legacy system—the platform can seamlessly insert a GUI agent, simulating human interactions with the software. This combination of API-based and GUI-driven actions makes it straightforward to build end-to-end automations, blending the speed and scalability of APIs with the accessible, visual nature of GUI-centric operations.


\subsection{Summary}
These strategies illustrate how the once-distinct boundaries between API-based and GUI-based agents are gradually merging. The hybrid approach harnesses the flexibility and universality of GUI-driven interfaces alongside the reliability and performance of direct API calls. This convergence allows for more comprehensive automation, catering to diverse scenarios that range from rapid data processing to intricate user-interface validation, meeting varied requirements, whether efficiency, user-centric validation, or rapid development. Although further research and refinement are needed, these converging paradigms foreshadow a future in which agent-based automation is both wide-ranging and intelligent, adapting seamlessly to the evolving complexities of modern software ecosystems.

\section{API vs.\ GUI Agents: Strategic Considerations}
\label{sec:strategic_considerations}

\begin{table*}[t]
\centering
\caption{Strategic criteria for selecting agent paradigms.}
\label{tab:strategic_considerations}
\begin{tabular}{p{4.5cm}| p{4.cm}| p{6.8cm}}
\hline\hline
\textbf{Scenario} & \textbf{Recommended Approach} & \textbf{Rationale} \\
\hline
Stable, well-documented APIs & API Agents & Exploit robust endpoints for speed and reliability \\\hline
Performance-critical operations & API Agents & Reduce latency and overhead via direct function calls \\\hline
Controlled access to applications & API Agents & Ensure  safety and security \\\hline
Legacy or proprietary software & GUI Agents & Automate tasks without requiring new backend integration \\\hline
Visual validation or UI testing & GUI Agents & Verify on-screen text or elements directly \\\hline
Interactive or graphical manipulation & GUI Agents & Seamlessly replicate human-like interactions with visual elements \\\hline
Partial API coverage & Hybrid & Combine UI-based steps where APIs are unavailable with direct calls for data-heavy tasks \\\hline
Future-proofing & Hybrid & Facilitate switching from GUI to API as endpoints evolve \\\hline
\hline
\end{tabular}
\end{table*}

While the preceding sections have contrasted API- and GUI-based LLM agents in terms of architecture, reliability, efficiency, and potential convergence, many practical deployments hinge on a more fundamental question: \emph{which paradigm should be employed under various real-world conditions?} This section provides guidance on selecting the most suitable strategy.

\subsection{When to Favor API Agents}
API-based agents tend to be the most compelling choice when well-defined programmatic interfaces exist. Official, stable APIs typically come with rigorous documentation and versioning, enabling strong error handling and consistent performance. In such an environment, developers can harness the inherent speed and reliability of API calls to execute tasks efficiently, thereby minimizing system overhead and latency. This approach is especially advisable when applications are designed for backend integrations, or when enterprise-level reliability is paramount for mission-critical workflows. By leveraging stable endpoints, API-based agents can also reduce long-term maintenance burdens, as changes in the system often entail versioned updates rather than complete interface overhauls.

In addition, API agents provide controlled access to applications, restricting functionality to a predefined and manageable scope. This is a crucial consideration in agent-based systems, where safety and security are paramount. In such cases, API agents are ideal, as their actions are confined to a constrained set of operations, ensuring predictable and secure interactions.

\subsection{When to Favor GUI Agents}
GUI-based agents become particularly relevant in scenarios where no direct API exists, or the available APIs provide only partial coverage of the required automation tasks. This is especially evident in mobile applications, where each app operates as an isolated environment, restricting external API access. Furthermore, system-level operations on mobile devices often require root access, further limiting API usability and necessitating GUI-based automation as an alternative.

Another key advantage of GUI agents is their ability to perform visual validation, which is essential in workflows that require confirming on-screen text, UI element positioning, or interface consistency before taking action. In such cases, a GUI agent, which interacts with software much like a human user, offers clear benefits over an API-based approach. Similarly, legacy or proprietary systems that lack extensible backend services can leverage GUI-driven automation, allowing agents to navigate existing interfaces without requiring modifications to the underlying codebase or the development of new APIs. This makes GUI agents particularly valuable in enterprise environments where implementing and maintaining new API integrations would be impractical or cost-prohibitive.

Moreover, GUI agents are inherently well-suited for applications that rely on interactive or graphical manipulation. Tasks such as creating animations, drawing in Photoshop, or interacting with complex design tools are best executed through direct visual interactions rather than API commands. In these cases, GUI-based automation closely mirrors the natural way humans interact with such applications, making it the preferred choice over an API-driven approach.

\subsection{When to Consider a Hybrid Approach}
A hybrid strategy combines the strengths of both paradigms, providing a unified workflow that can accommodate a wide range of requirements. This approach is particularly advantageous when some aspects of the task map neatly onto existing APIs, while other components remain exclusively accessible through a graphical interface. In such cases, using an API-based agent for data-intensive or programmatically streamlined operations preserves performance, while a GUI agent handles specialized front-end interactions or visual validations. Moreover, adopting a hybrid solution offers flexibility for future system evolution; as new APIs become available, tasks initially managed via the GUI can be seamlessly transitioned to API calls.

\subsection{Summary}
In summary, the strategic considerations for deploying API-based versus GUI-based agents depend on the nature of the target software, the level of integration or validation required, and long-term sustainability concerns. API agents excel when stable, documented endpoints exist, offering a reliable and performant mode of automation. GUI agents are advantageous in contexts where interfaces are the only means of access or where visual confirmation is essential. Finally, hybrid approaches strike a balance between these strengths, allowing organizations to adapt as their software ecosystems evolve. By taking these factors into account, decision-makers can ensure they select the agent paradigm that best aligns with their specific requirements.

\section{Conclusion \& Looking Forward} \label{sec:conclusion}
The advent of LLM-based agents represents a significant leap forward in automation. These agents embody two core paradigms: one centered on well-defined programmatic interfaces (API agents) and one rooted in human-like interactions with graphical interfaces (GUI agents). By design, these paradigms differ in their operational principles, leading to a divergence in architectural choices, performance profiles, and real-world applicability. However, they also exhibit complementary strengths—API agents excel at speed, security, and reliability, while GUI agents offer flexibility, broad applicability, and transparency—making them poised for a future of hybridization and convergence.

\textbf{Looking Ahead.} The ongoing maturation of LLM technologies will likely reinforce both strands of agent development. On one hand, increasingly capable coding assistants promise to simplify the creation and maintenance of APIs, thereby enhancing the scalability of API agents. On the other hand, the rise of powerful multimodal models will expand the scope of GUI-based agents, enabling more robust visual understanding and sophisticated manipulation of GUIs. These point to an evolving ecosystem in which API-centric and GUI-centric approaches become increasingly interwoven.

Looking forward, the seamless integration of these agent types may give rise to entirely new forms of software—tools that automatically generate or refine APIs for efficient back-end operations and also dynamically orchestrate user-interface elements for transparent front-end interactions. This confluence of paradigms has the potential to transform human-computer interaction, blurring the boundaries between what is generated by code and what is experienced through a visual interface. In the long term, it could reshape how we conceive of software development, user experience, and the broader workflows that underlie digital ecosystems.

\bibliography{custom}
\bibliographystyle{plainnat}

\appendix
\begin{table*}[t]
\centering
\caption{APIs supported across Office applications.}
\label{tab:office_apis}
    \begin{tabular}{l|c|l}
    \toprule
    \textbf{API} & \textbf{Application} & \textbf{Description} \\ \midrule
    \texttt{select\_text}          & Word       & Select matched text in the document. \\\hline
    \texttt{select\_paragraph}     & Word       & Select a paragraph in the document. \\\hline
    \texttt{set\_font}             & Word       & Set the font size and style of selected text. \\\hline
    \texttt{save\_as}              & Word       & Save the current document to a desired format. \\\hline
    \texttt{insert\_excel\_table}  & Excel      & Insert a table at the desired position. \\\hline
    \texttt{select\_table\_range}  & Excel      & Select a range within a table. \\\hline
    \texttt{reorder\_column}       & Excel      & Reorder columns of a table. \\\hline
    \texttt{save\_as}              & Excel      & Save the current sheet to a desired format. \\\hline
    \texttt{set\_background\_color}& PowerPoint & Set the background color of slide(s). \\\hline
    \texttt{save\_as}              & PowerPoint & Save the current presentation to a desired format. \\
    \bottomrule
    \end{tabular}
\end{table*}

\section{APIs for Office Applications}
\label{sec:office_apis}
To demonstrate the practicality of integrating API-level control into desktop automation, we developed a set of \textit{12} task-specific APIs across three popular Microsoft Office applications: Word, Excel, and PowerPoint. These APIs are designed to complement GUI-based interaction by enabling direct access to underlying application functions through COM interfaces.

For \textbf{Word}, the supported APIs include text and paragraph selection, font formatting, and file export. For \textbf{Excel}, the APIs provide capabilities such as table insertion, range selection, column reordering, and sheet export. For \textbf{PowerPoint}, we implement functions to change slide background colors and save presentations. Notably, a common \texttt{save\_as} API is reused across all three applications, illustrating the potential for cross-application generalization.

Table~\ref{tab:office_apis} summarizes the available APIs and their functionalities. These APIs serve as examples of how direct function invocation can improve reliability and reduce the number of GUI actions needed, particularly for tasks involving complex formatting or structured content manipulation.


\end{document}